\documentclass[10pt,conference]{IEEEtran}
\IEEEoverridecommandlockouts

\usepackage{amsmath,amssymb,amsfonts}
\usepackage{graphicx}
\usepackage{graphics}
\usepackage[printonlyused]{acronym}
\usepackage{xcolor}
\usepackage{glossaries}
\usepackage[sorting=none]{biblatex}
\begin{document}

\title{GAN-Based Content Generation of Maps for Strategy Games\\
\thanks{
Published in the Proceedings of GAME ON 2022; Cite as:\\
Nunes, V., Dias, J., Santos, Pedro A.: GAN-Based Content Generation of Maps for Strategy Games. Proceedings of GAME-ON'2022, pg 20-31, ISBN 978-9-492859-22-8}
}

\author{
\IEEEauthorblockN{Vasco Nunes}
\IEEEauthorblockA{\textit{Instituto Superior Técnico}\\
\textit{University of Lisbon} \\
Lisbon, Portugal \\
vasco.nunes@tecnico.ulisboa.pt}

\and

\IEEEauthorblockN{João Dias}
\IEEEauthorblockA{\textit{Faculty of Science and Technology} \\
\textit{University of Algarve and CCMAR and INESC-ID}\\
Faro, Portugal \\
jmdias@ualg.pt}

\and

\IEEEauthorblockN{Pedro A. Santos}
\IEEEauthorblockA{\textit{Instituto Superior Técnico / INESC-ID}\\
\textit{University of Lisbon} \\
Lisbon, Portugal \\
pedro.santos@tecnico.ulisboa.pt}
}

\maketitle


\begin{abstract}
Maps are a very important component of strategy games, and a time-consuming task if done by hand.  Maps generated by traditional PCG techniques such as Perlin noise or tile-based PCG techniques look unnatural and unappealing, thus not providing the best user experience for the players. However it is possible to have a generator that can create realistic and natural images of maps, given that it is trained how to do so. We propose a model for the generation of maps based on Generative Adversarial Networks (GAN). In our implementation we tested out different variants of GAN-based networks on a dataset of heightmaps. We conducted extensive empirical evaluation to determine the advantages and properties of each approach. The results obtained are promising, showing that it is indeed possible to generate realistic looking maps using this type of approach.
\end{abstract}

\section{Introduction}\label{intro}

In maps for strategy games, the map's visual characteristics play a very important role in the player's experience. When we talk about visual characteristics, we usually refer to the map's outline and level of detail. Complex elements like peninsulas, mountain ranges or islands provide more tactical information, improving the player's decisions in a strategic way. Therefore, these details improve the way the information contained in the map is assessed, in order for the player to make decisions in a strategic way. A game which uses the same kind of map numerous times with no variety can cause players to become bored after replaying the game a few times.

One of the ways to generate maps for strategy games is by using Procedural Content Generation (PCG) \footnote{Creation of game content algorithmically with limited or indirect user input} techniques. The most common traditional approach for initial Heightmap generation is the Perlin Noise \cite{10.1145/325165.325247}, followed by complementing techniques such as Hydraulic erosion \cite{Mei2007}.
Unfortunately the maps generated look a bit unnatural and unappealing. 
More so, most of the methods generated by PCG suffer from some kind of uncontrollability. The ideal scenario would be to have a generator that learned 
to create realistic images of maps and with the complex elements (peninsulas or mountain ranges) appearing next to each other\footnote{From an interview conducted for this research with Andy Gainey, gameplay programmer at Paradox Development Studio}.



Recent advances in Deep Neural Networks, and in particular GANs (Generative Adversarial Networks) highlight the potential for a new approach for the automatic generation of maps. GAN is a framework proposed by Ian J. Goodfellow et al. \cite{Goodfellow2014} that is trained to generate data (images in most cases) with the same characteristics of a given training set. For example, a GAN trained on images of human faces would be able to generate realistic samples that are authentic to human observers.

The framework consists of two networks competing against each other, thus the term adversarial: a generative network, Generator $G$, which creates fake data from a random distribution $p_{z}$ (usually normal or uniform) and a discriminative network, Discriminator $D$, which, by giving it some data $x$, estimates whether $x$ came from real data distribution $p_{data}$ or from the generator's distribution $p_{g}$.

A real world analogy would be the job of an art counterfeiter and a cop. The cop ($D$) learns to detect false paintings while the counterfeiter ($G$) improves on producing perfectly fake paintings indistinguishable from real ones.

\begin{figure}[h]
\centering
\includegraphics[width=0.45\textwidth]{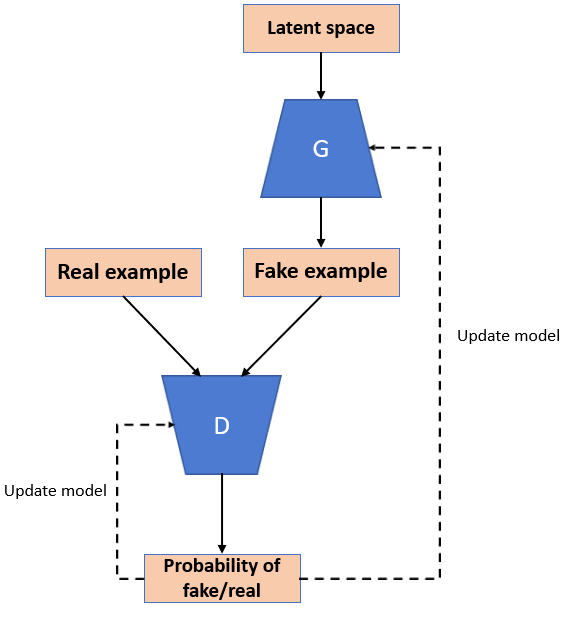}
\caption{Generative Adversarial Network Model architecture.} 
\label{fig:GAN}
\end{figure}

The generation of images using GANs reached great success in recently years. Recent applications using this network include creation of realistic faces \cite{Karras2018}, pose guided person image generation \cite{Mirza2018} or transforming images from one domain to another \cite{Zhu2017}.

Taking this into account, the research problem addressed in this work is to explore several GAN's techniques in order to generate realistic and appealing maps for strategy games.

By ``realistic", we mean that maps should be perceived in a similar way to natural land formations. By ``appealing", we mean that maps should have suitable characteristics and elements discussed above, that allow players to have a more challenging and interesting experience.

Taking into account the research problem, if we have a proper and balanced heightmap's dataset of natural landscapes, we believe that this type of technique can be successfully used to generate maps that resemble the original dataset, which will have the type of characteristics existing in natural landscapes, such as peninsulas and mountain ranges. Starting from a baseline GAN architecture, we will slowly improve its structure in order to achieve our goal, taking into account the proper evaluation of the models.


\section{Related Work}

In this section we describe different types of GANs that appeared in recent works and that we will use.


\subsection{DCGAN}\label{DCGAN}

In the work of Alec Radford et al. \cite{Radford2016} the authors managed to consolidate the junction of GAN and Convolutional Neural Network (CNN) frameworks, after several unsuccessfully attempts to do so in the past years. 

CNN are a subset of neural networks most commonly applied to image and video recognition or computer vision problems. They were largely inspired by the visual cortex, small regions of cells sensitive to specific regions of the visual field \cite{DeepLearning}. These networks are mostly composed by convolutional layers, that are responsible for applying the convolution\footnote{In image processing, refers to the process of adding each pixel of the image to its local neighbors, weighted by a kernel.} operation of the input using a filter, or kernel, and sending the result, also known as feature map to the next layer. If this kernel is designed to detect a specific type of feature on the input, then filtering it across the whole image would allow the kernel to detect that feature anywhere in the image independently of the feature's location. By having this translation invariance characteristic and a shared-weight architecture, CNN are also known as Shift Invariant Artificial Neural Networks \cite{Zhang1990}.

Their methodology consisted in adopting and modifying three demonstrated changes to CNN architectures:
\begin{enumerate}
    \item Replacing the pooling layers of the CNN baseline architecture for convolutional layers with stride 2 
    in order for $G$ and $D$ to learn its own spatial upsampling and downsampling respectively;
    \item Elimination of fully connected layers for convolutional layers. To make the most use of these convolution layers, they added another layer at the beginning of $G$ that takes the input vector $z$ and reshapes it into a 4-dimensional tensor. They also added another layer at the end of $D$ that flattens the image to a single value output;
    \item Applying Batch Normalization to layers in order to stabilize learning by normalizing the input of a layer to have zero mean and unit variance, for each minibatch. 
\end{enumerate}


\subsection{Progressively Growing GANs} \label{PGGAN}

Generation of high-resolution images is a difficult task since it is easier to discriminate between the fake and real images. In the work of Karras et al. \cite{Karras2018}, the authors proposed a training methodology that consists on starting with low-resolution images, and then progressively increasing the resolution by adding layers to both the discriminator and generator (which are mirror images of each other and always grow in synchrony). In other words, the authors are slicing a bigger complex problem into smaller ones and slowly increasing the complexity to prevent the training from becoming unstable. The incremental addition of the layers allows the models to effectively learn coarse-level detail and later learn even finer detail, both on the generator's and discriminator's side.

The insertion of layers cannot be done directly due to sudden shocks to the already well-trained layers. Instead they phase in the layer. This operation consists on using a skip connection\footnote{Skip connections are connections of outputs from early layers to later layers through addition or concatenation} to connect the new block to the input of $D$ or output of $G$ using a weighted sum with the existing input or output layer, that represents the influence of the new block. It is controlled by a parameter $\alpha$ that starts at a very small value and increases linearly to 1 over the process of training. In other words, we can think of this operation as the layers slowly being inserted during the training phase. 

In terms of results, this network was capable of generating high resolution images of $1024 \times 1024$, creating a high-resolution version of the CELEBA dataset \cite{liu2015faceattributes}.




\subsection{Wasserstein GAN} \label{WassGAN}


Martin Arjovsky et al. \cite{arjovsky2017wasserstein} propose a GAN that has an alternate way of training so that the generator model better approximates the distribution of data of a given dataset. More so, they present an alternate loss function in which $D$ is always giving enough information for the $G$ to improve himself, even if $D$ has reached its optimality.

The discriminator $D$ is replaced by a critic $C$ that instead of classifying an image as fake or real (in the interval [0,1]), it scores the fakeness or realness of an image (in the interval ]$-\infty$,$+\infty$[). 
This score is also known as Wasserstein estimate. The critic is looking to estimate the Wasserstein distance between the dataset sample distribution and the generated images distribution, which corresponds to the distance between the average critic score on real and the average critic score on fake images. Thus the network's objective function can be summarized as follows:
\begin{itemize}
    \item $C$ objective function is the difference between the average critic score on fake images and the average critic score on real images;
    \item $G$ objective function is the average critic score on fake images.
\end{itemize}

Both the networks are trying to maximize these objective functions. Therefore, for $G$, a larger score of the fake images will result in a higher output for the $G$, encouraging $C$ to output higher scores for the fake images.
For $C$, a larger score for real images results in a lower value for the model, penalizing it, thus the encouragement for the critic to score lower scores for the real images.

\subsection{VAE + GAN} \label{VAEGAN}

The idea of combining the power of Variational Autoencoders (VAE) and GAN was the basis for the work of Larsen et al. \cite{larsen2016autoencoding}.  The motivation behind their work was to leverage learned representations to better measure similarities in the data distribution. 

Autoencoders are another subset of neural networks used to generate images with good results. They learn a representation of a given data \cite{Bengio2009}. They are commonly used for face recognition or acquiring semantic meanings of words. The general idea behind this neural network is that they learn to copy the input to the output through a latent space that compresses the input maintaining only the most relevant and important information. 
The decoder then decompresses the information retained in the latent space ( also known as code) leading to a very similar copy of the output.

VAE are a subset of autoencoders specialized in content generation. They inherit the architecture from the traditional autoencoders but instead of mapping the input to a fixed latent space, the VAE maps the input onto a latent distribution, which allows us to take random samples from the latent space. These samples are then decoded using the decoder segment to generate outputs very similar to the inputs used to train the encoders. Instead of having a fixed vector as the latent space, the vector is replaced by two separate vectors, that represent the mean and the standard deviation of the distribution, respectively.

Typically, VAE uses element-wise similarity in their reconstruction error, 
however, Larsen et al. \cite{larsen2016autoencoding} propose using the GAN discriminator to measure the sample similarity. In other words, they use the GAN's discriminator as a way to measure the difference between the VAE's output and the original image. 


The authors' proposed architecture is comprised of a single model that simultaneously  learns to encode, generate and compare samples from the dataset. The GAN's generator coincides with the VAE's decoder. The authors define the loss of the model as:
\begin{equation}
    \mathcal{L} = 
    \mathcal{L}_{VAE} + \mathcal{L}_{GAN}
\end{equation}
where $\mathcal{L}_{GAN}$ is the Binary cross entropy loss and $\mathcal{L}_{VAE}$ is defined as:
\begin{equation}
    \mathcal{L}_{VAE} = \mathcal{L}_{prior} + \mathcal{L}^{Dis_{l}}_{llike} 
\end{equation}
where
$\mathcal{L}_{prior}$ is a prior regularization term, the Kullback-Leibler divergence and $\mathcal{L}^{Dis_{l}}_{llike}$ is the expected log
likelihood (reconstruction error) expressed in the GAN discriminator:
\begin{equation}
    \mathcal{L}^{Dis_{l}}_{llike} = - \mathbb{E}_{z \sim Enc(x)} [\log p(Dis_{l}(x)|z)]
\end{equation}
with
$Dis_{l}$ denoting a hidden representation of the $l$th layer of the Discriminator.




\section{Dataset Creation}\label{Dataset creation}
In order to apply the GAN-based techniques, it is necessary to have a group of examples of what the system is supposed to generate. Therefore, we began by building a proper dataset of natural landscapes which had the characteristics of the already existing land formations of the planet Earth, such as peninsulas and mountain ranges. 


\subsection*{Data gathering}

\begin{figure}[h]\centering\includegraphics[width=0.37\textwidth]{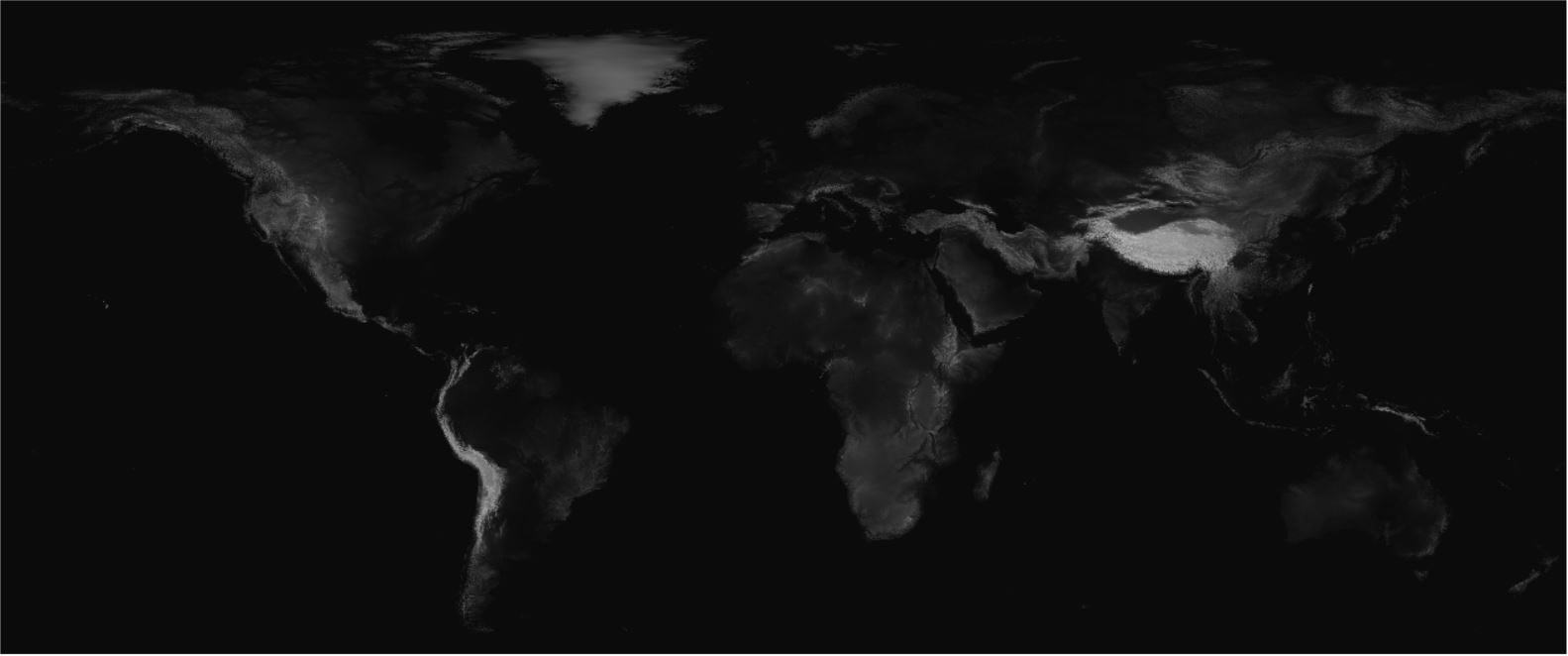}\caption{Heightmap of the planet Earth after joining the DEM files together}\label{fig:HeightMapEarth}\end{figure}

When looking for a proper dataset we had the idea that no data could resemble more the characteristics discussed in Section \ref{intro} than data from the real world, in which the terrain had already the desired complex elements and characteristics.
Taking this idea into account, we used a public dataset \footnote{https://www2.jpl.nasa.gov/srtm/cbanddataproducts.html} with nearly global coverage of the planet Earth generated by a satellite radar topography mission. The dataset consists of several Digital Elevation Model (DEM) files created by a Ground Data Processing System supercomputer with a 3 arc-second sample spacing. We grouped the different DEM files together and exported them into a Tagged Image File Format using a program called Global Mapper, resulting in a Heightmap image with $43200 \times 18000$ resolution, as shown on Fig \ref{fig:HeightMapEarth}. 

\subsection*{Preprocessing}

As one can see in Fig \ref{fig:HeightMapEarth} most regions of the Earth are a bit too dark making the terrain not very noticeable. This is not ideal for the techniques to be used since the altitude data should be more evenly distributed. In other words, we needed a higher range of pixel values so that their difference would be better distributed between $0$ and $255$, instead of the original linear mapping between altitudes and pixel values. So, using the same program, we changed the image's brightness level so that lower altitude regions could have higher pixel greyscale values.



We proceeded to crop the high resolution image into $1024 \times 1024$ images with a sliding window of $512$ pixels, which gave in total $2822$ images. 

\subsection*{Dataset Augmentation / Removal of unwanted images}

Due to the lacking of enough images on the dataset for a problem with a moderate level of complexity such as the one approached here, we had to perform dataset augmentation in order to increase the number of images. The procedure is described below:
\begin{enumerate}
    \item To the $43200 \times 18000$ image, we applied different sets of image processing techniques: rotation between $0$ and $180$ degrees, and horizontal / vertical flipping. The rest of the image would then be filled with the pixel value $255$.
    \item As done in the preprocessing phase, the $43200 \times 18000$ was cropped into $1024 \times 1024$ images with a sliding window of $512$.
    \item Resulting images that had little to no continental land or were cut due to the image processing techniques, were removed. In other words, images that had $95\%$ of the pixel values $\leq 25$, or had the pixel value of $255$ were removed.
\end{enumerate}

This procedure was repeated $15$ times, ending up with $12640$ images of $1024 \times 1024$ resolution, which we found more than reasonable for the problem. Unfortunately, Working with $1024 \times 1024$-sized images would result in our models having a high number of parameters thus requiring more computational resources, such as memory, which we had not at our disposal. Therefore, we had to downsize the dataset to a $128 \times 128$ resolution. The image rescaling was done using a nearest neighbor interpolation filter.





\section{GANs' Architecture and Training}


Instead of just focusing on one type of GAN, we decided to explore several models in order to decide which one would be the most appropriate for the problem, making a comparison
in terms of quality of the images generated, training efficiency and ease in training and convergence.
We added tables which detail each model's architecture in the Appendix.




\subsection{DCGAN}\label{chap3:DCGAN}


The Deep Convolutional Generative Adversarial Network (DCGAN) architecture was the first to be tested, to serve as an initial baseline model and a foundation for the other models. We followed the majority of the architectural guidelines explained in \cite{Radford2016} such as:
\begin{itemize}
    \item Using strided convolutions on $D$ and fractional-strided (transposed) convolutions on $G$, allowing the network its own spatial downsampling and upsampling;
    \item Batch Normalization layers in both networks, which helps the network on its learning process;
    \item Removal of fully connected layers and replacing them with the convolutional layers except in the beginning of $G$ and in the end of $D$;
    \item Leaky ReLU activation function for all layers of $D$, since it is more effective for models generating images with higher resolution;
\end{itemize}
However, instead of applying the Rectified Linear Unit (ReLU) activation function to all layers of $G$, we instead used the Leaky ReLU, since the latter activation function has been proven to be more effective than the ReLU \cite{wilson2016proceedings}. 

\begin{figure}[h]
\centering
\includegraphics[width=0.55\textwidth]{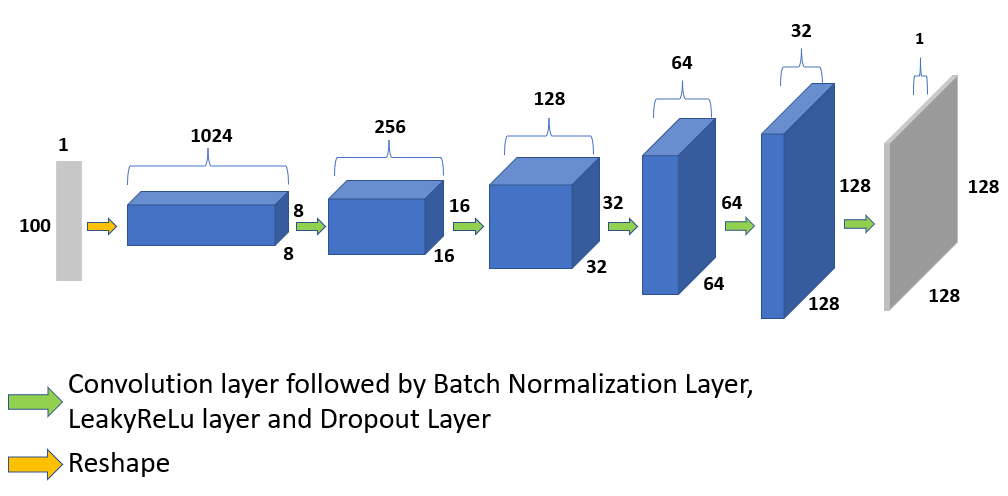}
\caption{DCGAN's Generator Model} 
\label{fig:DCGAN_G}
\end{figure}

\begin{figure}[h]
\centering
\includegraphics[width=0.55\textwidth]{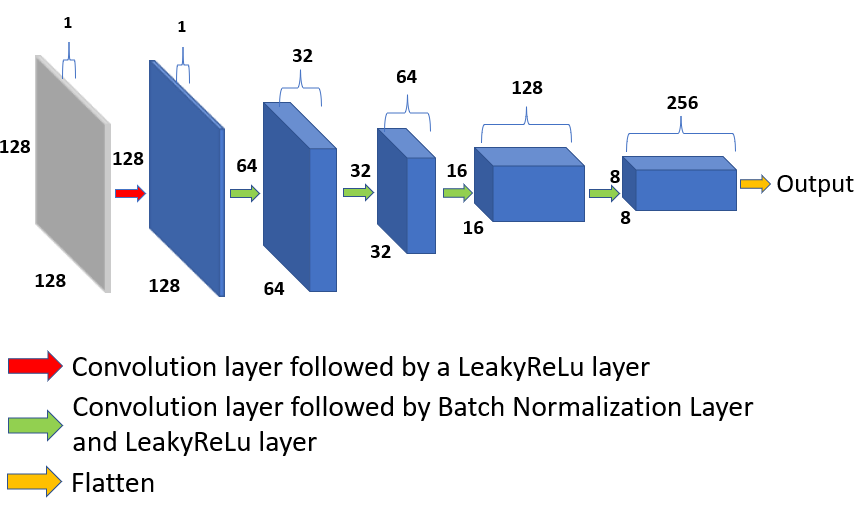}
\caption{DCGAN's Discriminator Model} 
\label{fig:DCGAN_D}
\end{figure}

Fig. \ref{fig:DCGAN_G} and Table \ref{fig:DCGAN_D} depicts the model of the GAN generator and discriminator, respectively. 

All of the convolutional and transposed convolution layers had a kernel size of $5$, SAME padding and a stride of $2$, except for the first convolution layer of $D$ which had stride of $1$. Those layers' weights were initialized using a zero mean normal distribution with standard deviation $0.02$. The values used for the Batch Normalization, Leaky ReLU layers and the kernel size on both networks were the same as the ones used in \cite{Radford2016}.

It is well-known that GANs suffer from the vanishing gradient problem, where an optimal $D$ ( one that is really good at classifying the images as real and fake), doesn't provide enough information for the $G$ to improve itself \cite{arjovsky2017wasserstein}, \cite{arjovsky2017principled}.

Regarding the DCGAN's training phase, in order to battle the GAN's vanishing gradient problem we decided to add three hindering methods that would help in achieving this goal. These hindering methods, 
have the purpose of preventing $D$ from rapidly reaching a perfect discriminator scenario \cite{Salimans2016}, \cite{isola2018imagetoimage}, \cite{arjovsky2017principled}.
\begin{itemize}
    \item Adding a unique noise vector to each sample of the minibatch of $m$ real samples from $p_{data}$ and the minibatch of $m$ fake samples;
    \item Applying One-sided Label Smoothing with $\beta = 0.2$. 
    \item Adding a Dropout layer after each Leaky ReLU layer on $D$. For these layers we chose to set 50\% of the input's values to $0$;
\end{itemize} 


We came up with three expressions for the $noise\_factor$. The first one illustrates an increasing $noise\_factor$ over the number of epochs:

\begin{equation}
\label{noise_1}
    noise\_factor = \frac{epoch \times 0.5}{epochs}
\end{equation}

In the second and third one, the $noise\_factor$ increases till half the number of epochs and then decreases.

\begin{equation}
\label{noise_2}
    noise\_factor = 
        \begin{cases} 
            \frac{ep \times 0.5}{half\_ep} &\quad \text{if } ep \leq half\_ep\\
            0.5 - \frac{ep \times 0.5}{ep} &\quad \text{otherwise}
        \end{cases}
\end{equation}

\begin{equation}
\label{noise_3}
    noise\_factor = 
        \begin{cases} 
            \frac{e \times 0.5}{h\_ep} &\quad \text{if } e \leq h\_ep\\
            0.5 - \frac{(e - h\_ep) \times 0.5}{h\_ep} &\quad \text{otherwise }
        \end{cases}
\end{equation}

In \eqref{noise_2} the noise drops to half after half of the epochs have passed (e.g. $0.3$ - $0.4$ - $0.5$ - $0.25$ - ...) while in \eqref{noise_3} the noise ascends and descends at a same rate (e.g $0.3$ - $0.4$ - $0.5$ - $0.4$ - $0.3$ - ...). We wanted to test what would happen if we hindered $D$ till half of the training phase and then ease it either by suddenly decreasing the $noise\_factor$ to half at the middle of the training phase or having a similar rate for increasing and decreasing the $noise\_factor$.



\subsection{WGAN}



The second tested architecture was the Wasserstein Generative Adversarial Network (WGAN) discussed in Section \ref{WassGAN}. With this model we were expecting to work on some of the problems that came with the DCGAN model such as the failure to converge and the vanishing gradient problems.


The structure of the generator from the DCGAN model and the WGAN model is exactly the same. The same goes for the discriminator from DCGAN model and the critic from WGAN model except we're using a linear activation function as the last layer on the critic instead of the sigmoid, since the score for realness or fakeness of an image doesn't have a limit value.

There are some things to be taken in consideration about the training phase:
\begin{itemize}
    \item As discussed in section \ref{WassGAN}, during each epoch, $C$ is trained $n\_critic$ times more than $G$;
    \item It was not to possible to implement the Wasserstein distance with the formulation presented in \cite{arjovsky2017wasserstein} using Keras' API  methods because Keras doesn't allow a sum of losses from independent batches. We used the fact that maximizing the Wasserstein distance is equivalent to increasing the distance between the Wasserstein score for positive examples and the score for negative examples. One simple way of achieving this is by returning positive estimates for real examples and negative estimates for fake examples\footnote{https://machinelearningmastery.com/how-to-implement-wasserstein-loss-for-generative-adversarial-networks/}.
    
    \item When updating the weights, we clip them to stay in an interval between a constant $-c$ and $c$. This clipping is necessary to ensure the critic’s approximation to the Wasserstein distance, as explained in \cite{arjovsky2017wasserstein}.
\end{itemize}

\subsection{ProgGAN} \label{ProgGAN}

For a third architecture we decided to apply the principles of the WGAN model and the idea of generating high-resolution images discussed in Section \ref{PGGAN} together to create a GAN
based on the Progressive Growing Generative Adversarial Network (ProgGAN). We wanted to verify if this technique would have interesting results in terms of training time optimization.  We also wanted to see if it would generate images with higher resolution while maintaining a good quality. 

\begin{figure}[h]
\centering
\includegraphics[width=0.5\textwidth]{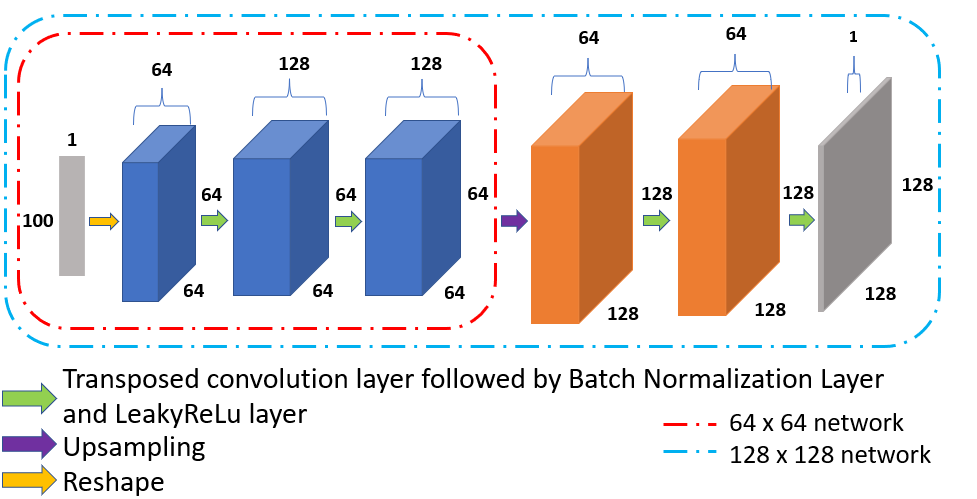}
\caption{ProgGAN's Generator Model} 
\label{fig:ProgGAN_G}
\end{figure}

\begin{figure}[h]
\centering
\includegraphics[width=0.5\textwidth]{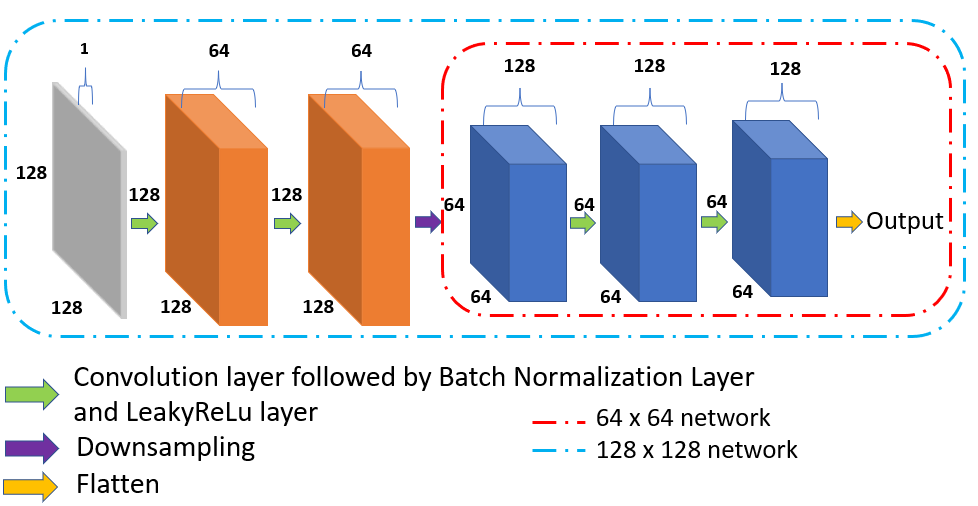}
\caption{ProgGAN's Critic Model} 
\label{fig:ProgGAN_D}
\end{figure}


Fig. \ref{fig:ProgGAN_G} and \ref{fig:ProgGAN_D} depict the models of the ProgGAN's generator and critic, respectively.

The upsampling layers use an upscaling factor of $2$ for both dimensions and a nearest neighbor interpolation. The downsampling layers perform an average pooling operation, using a downscaling factor of $2$ for both dimensions and a stride of $2$. During training, the phasing in from giving full weight to the $64 \times 64$ model (at the beginning of the training phase) to full weight to the $128 \times 128$ model (end of the training phase) is done through a weighted sum layer controlling how much to weight the input from the $64 \times 64$ and the $128 \times 128$ models. It uses a parameter $\alpha$ that grows linearly over the training phase. 

Each of the models was trained separately, in this order: $64 \times 64$ model $\rightarrow$ growth model $\rightarrow$ $128 \times 128$ model. We treated each of the models as being a WGAN.

\subsection{VAE + WGAN}

As a final contribution, we wanted to combine the VAE and WGAN together. The idea was to train a VAE to encode and decode the images of our dataset of heightmaps. We hoped that using a generator that was already trained to create features from our dataset, such as in the VAE, would accelerate the WGAN's training by increasing the efficiency in training time. 

\begin{figure}[h]
\centering
\includegraphics[width=0.5\textwidth]{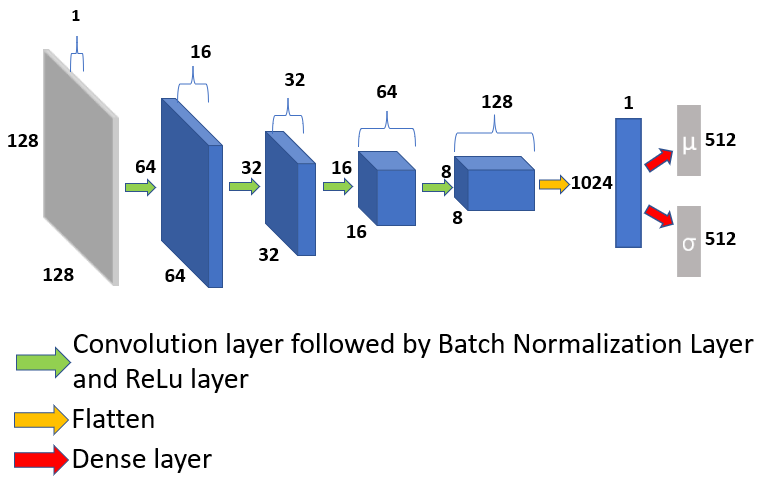}
\caption{VAE + WGAN's Encoder Model} 
\label{fig:VAE}
\end{figure}

Fig. \ref{fig:VAE} depicts the model of the encoder.

\begin{figure*}[h!]
\centering
\includegraphics[width=0.8\textwidth]{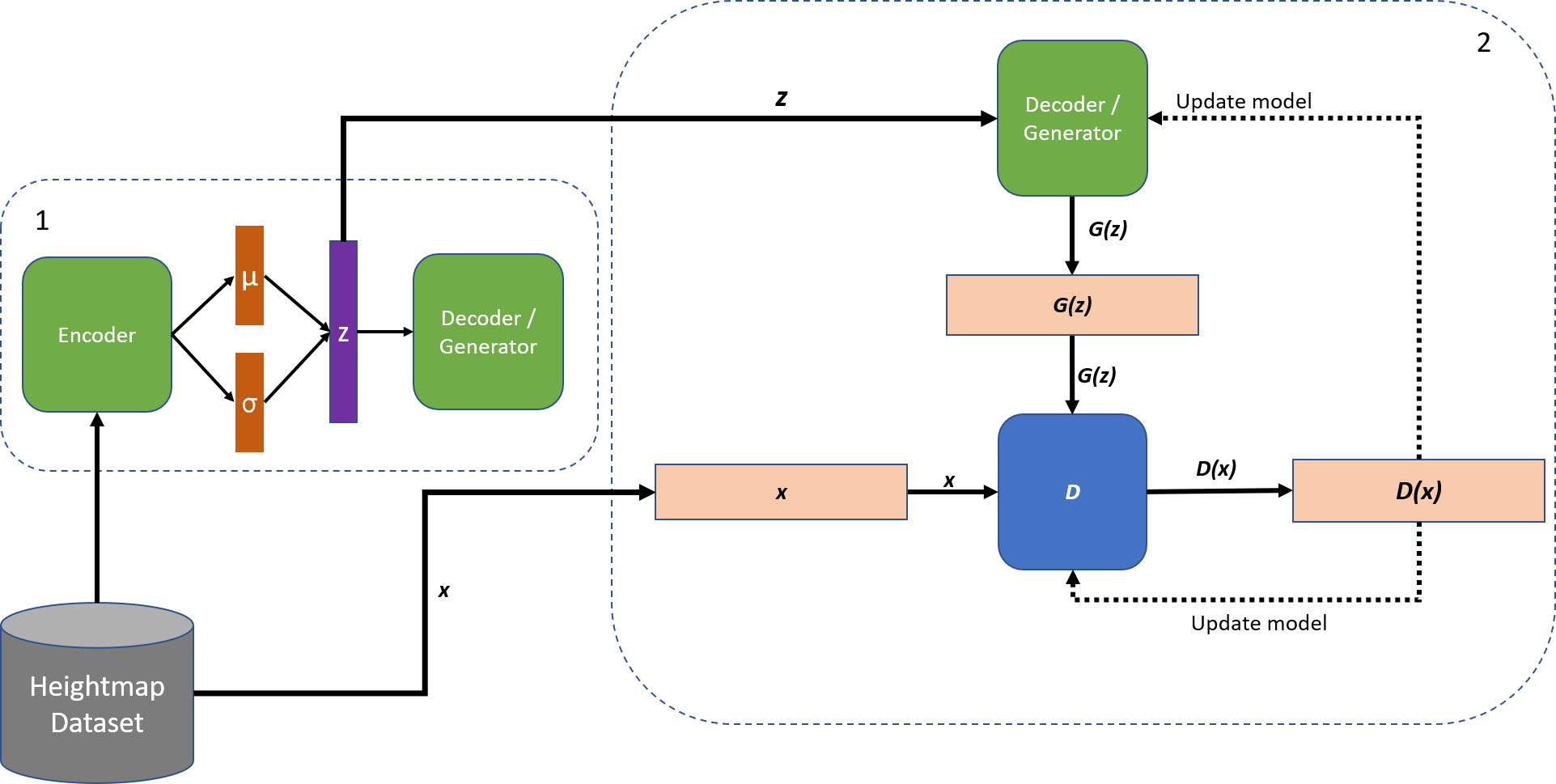}
\caption{VAE + WGAN architecture. Train the VAE (1) to learn the representation of the Heightmap dataset. Then use the sample vector $z$ created from the distribution's mean and standard deviation to train the WGAN (2). The VAE's decoder will be the generator of the WGAN.}
    \label{fig:VAEGAN}
\end{figure*}

The decoder and the discriminator followed the same structure as the WGAN's generator and critic. The only difference is that the last layer of the decoder uses a sigmoid activation function instead of the hyperbolic tangent used in WGAN's generator.

In terms of training, we trained the VAE and WGAN separately. We started with the VAE, by just training the model for a fixed number of epochs. Then we took the decoder model of the VAE and used it as our WGAN's generator. 
However, instead of normalizing our dataset between $-1$ and $1$ for both the networks, we normalized it between $0$ and $1$ since the last activation function of the decoder is a sigmoid, which ranges between the latter values. For some of the experiments, instead of generating the $z\_vectors$ from $\mathcal{N}(\mu,\,\sigma^{2})$\, we generated them using the vectors $\mu$ and $\sigma$ learned from the VAE. With this latter idea, we wanted to test if the decoder would be able to generate images with closer resemblance to the Heightmap dataset, since the vectors $\mu$ and $\sigma$ contained features of that dataset.











\section{Results}

All experiments were done in a desktop workstation with architecture x86\_64, CPU AMD Ryzen 5 2600X Six-Core processor and one GeForce RTX 2080 Ti graphic cards with 11 GB RAM. For the implementation and training of the networks we used Keras\footnote{https://keras.io/, the version used was 2.3.1}, a deep learning API running on top of 
 Tensorflow\footnote{https://www.tensorflow.org/, the version used was 1.14.0}. It provides abstractions and building blocks for developing and solving machine learning solutions. 

\subsection{DCGAN}


In terms of results, in the four experiments we performed, the networks were compiled 
using the Adam optimizer 
with learning rate $0.0002$ and $\beta_1 \footnote{Exponential decay for the running average of the gradient}  = 0.5$. Both values were chosen as suggested in \cite{Radford2016}.
The networks were trained for 1000 Epochs.
A brief description of every experiment made is described below:
\begin{itemize}
    \item E1: None of the discriminator hindering methods were used;
    \item E2: All the discriminator hindering methods were implemented; Equation \eqref{noise_1} for the noise vector;
    \item E3: All the discriminator hindering methods were implemented; Equation \eqref{noise_2} for the noise vector; 
    \item E4: All the discriminator hindering methods were implemented; Equation \eqref{noise_3} for the noise vector; 
\end{itemize}



\begin{figure}[h]
\centering
\includegraphics[width=0.5\textwidth]{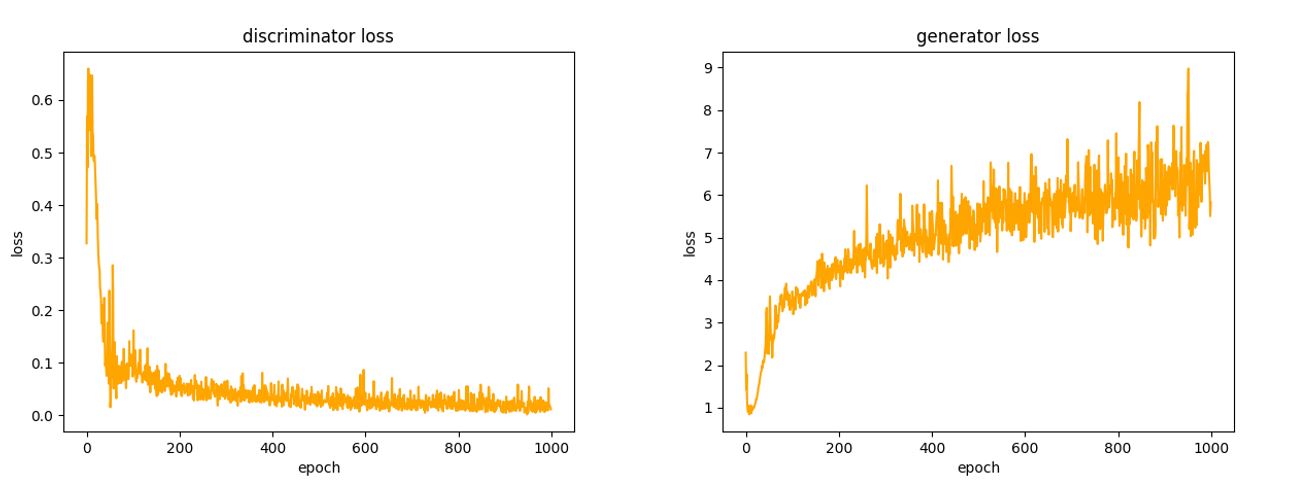}
\caption{$D$ and $G$ Binary Cross Entropy loss of the Experiment $1$}
\label{fig:DCGAN_graphics_1}
\end{figure}

In Fig. \ref{fig:DCGAN_graphics_1} one can observe the losses of $D$ and $G$ throughout the epochs for E1, and we can see the GAN's vanishing gradient problem appearing.

As we can observe, $D$'s loss rapidly converges to $0$, while $G$'s loss keeps growing unsteadily. Early during the training phase, $D$ reaches its optimal state, which is perfectly discriminating between the real and fake images. Thus, $G$ is unable to improve, leading to a loss growth.

For E2, E3 and E4, despite the fact that the hindering methods above mentioned did in fact help battle the GAN's vanishing gradient problem, the results obtained have a poor quality.

\begin{figure}[h!]
\centering
\includegraphics[width=0.45\textwidth]{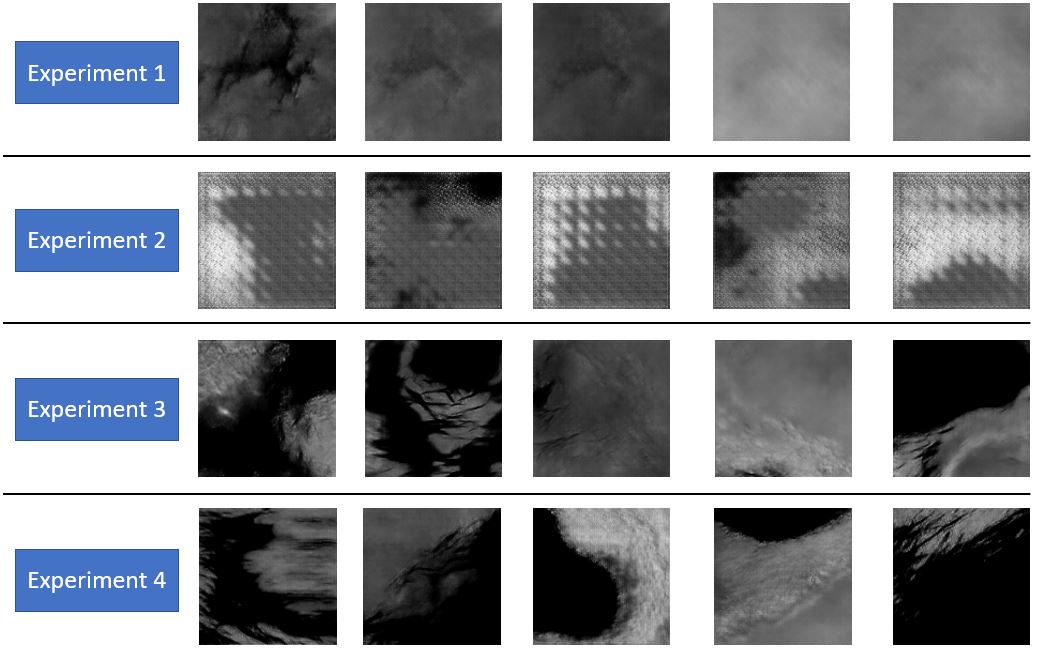}
\caption{Some of the images generated by DCGAN's generator after training for $1000$ epochs, in the several experiments.}
\label{fig:DCGAN_final_experiment_images}
\end{figure}




In E1, we can see that the model entered in a mode collapse, since the images generated have similar shapes, with some of them even having a blurry artifact. In E2 however, the quality of the images decayed, with the resulting images presenting some kind of artifacts in the form of a squared pattern. Experiments E3 and E4 yielded better results in terms of quality. Despite this results, all the images represented in Fig. \ref{fig:DCGAN_final_experiment_images} were classified as fake by the experiment's respective discriminators. 



\subsection{WGAN}

The network was compiled using the RMSProp optimizer with learning rate of $0.0005$, as in \cite{arjovsky2017wasserstein}. In terms of results, we first ran some tests to see which value we should use as a constraint for clipping the weights values from this list of values: $0.01$, $0.02$, $0.05$, $0.1$, $0.15$, $0.2$. 
We ended up choosing $0.1$ as the constraint since it gave the best results in terms of the Wasserstein estimate. 

\begin{figure}[h]
\centering
\includegraphics[width=0.5\textwidth]{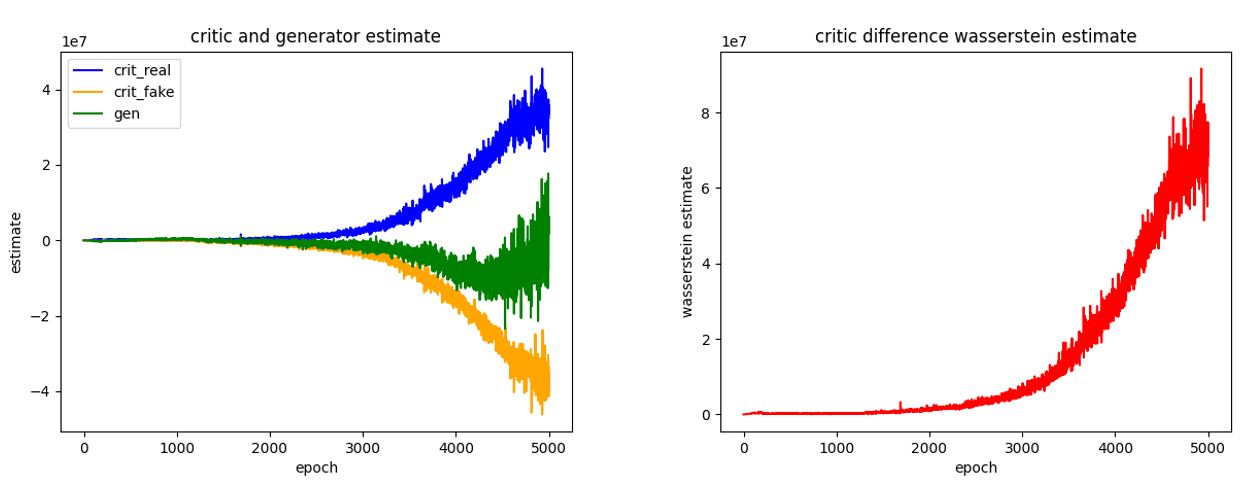}
\caption{Left: Wasserstein estimate for $C$ real and fake images and $G$ generated images during the training phase; Right: Difference between the $C$ Wasserstein estimate of the real images and the fake images}
\label{fig:WGAN_final_experiment}
\end{figure}

We ran a single but extensive experiment E5 where we trained the network for $5000$ epochs with a training time of $35$ hours and $27$ minutes. Fig. \ref{fig:WGAN_final_experiment} depicts the Wasserstein estimates of E5. The blue and yellow lines, which represent the Wasserstein estimate for the critic real and fake images, respectively, are distancing from each other. We can also verify this by looking at the other graphic, where the red line, which represents the difference between those two estimates, keeps growing troughout the training epochs.

\begin{figure}[h]
\centering
\includegraphics[width=0.45\textwidth]{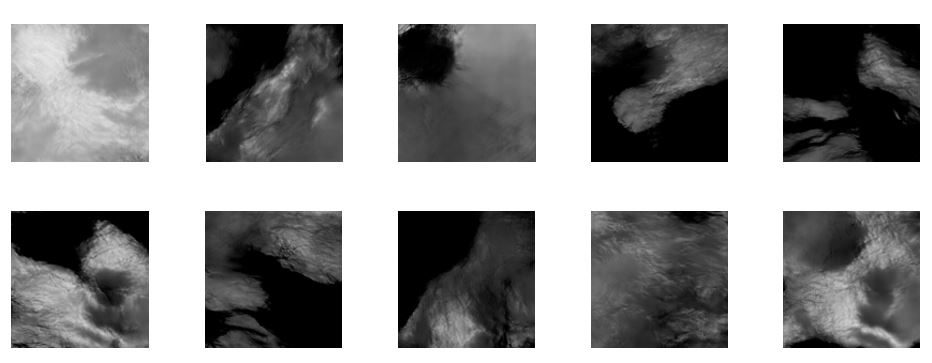}
\caption{Images generated by WGAN's generator during the last $5$ training epochs}
\label{fig:WGAN_final_experiment_images}
\end{figure}

\begin{figure}[h]
\centering
\includegraphics[width=0.30\textwidth]{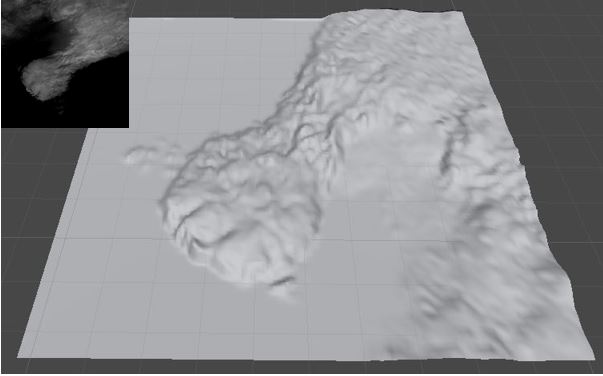}
\caption{3D representation of a heightmap generated in E5}
\label{fig:WGAN_3D}
\end{figure}

Fig. \ref{fig:WGAN_final_experiment_images} depicts some of the images generated during the last epochs. The generator was able to reproduce realistic images with a relative good quality and level of detail. Some complex structures are also present such as peninsulas, mountain ranges and groups of islands. Fig. \ref{fig:WGAN_3D} depicts a 3D representation of a heightmap generated in E5, where we can see a peninsula and some small islands next to it.

\begin{figure}[h!]
\centering
\includegraphics[width=0.45\textwidth]{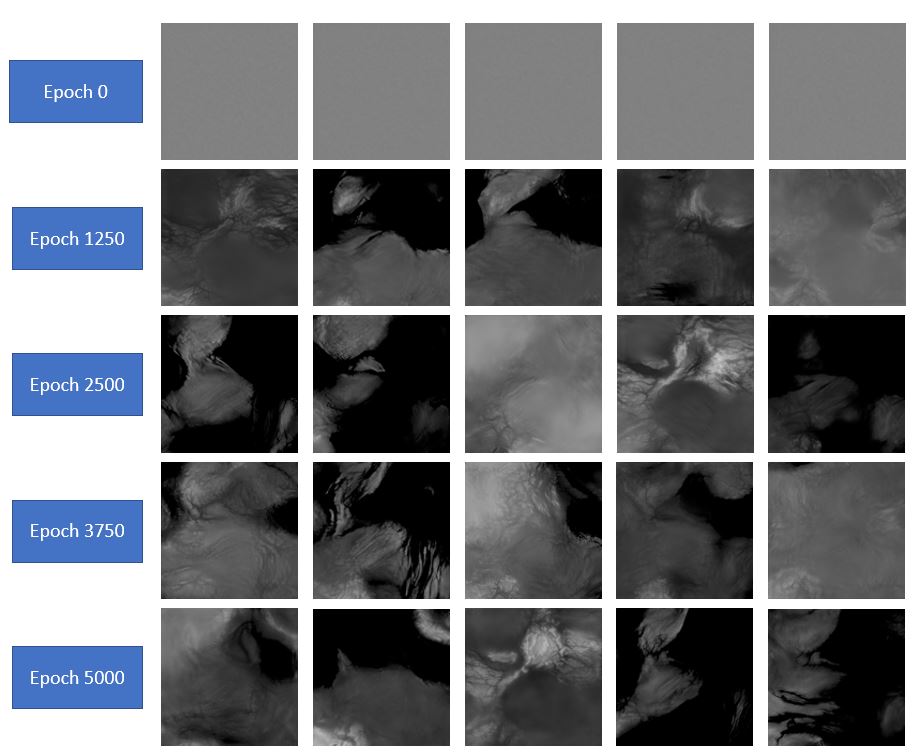}
\caption{Images generated by the $128 \times 128$ model's generator during the training of E6}
\label{fig:ProgGAN_images_128x128_only}
\end{figure}

Given the results we achieved with the previous experiment, we decided to run another experiment in order to check what type of results another WGAN, with a more complex structure and higher number of neurons than the previous one, could generate. So, in experiment E6 we trained the $128 \times 128$ model presented in subsection \ref{ProgGAN} for $5000$ epochs, which took 217 hours of training time. Fig. \ref{fig:ProgGAN_images_128x128_only} shows some of the images generated through some training epochs of E6. Using a WGAN with a more complex structure proved not to be so efficient, given that the images between experiments E5 and E6 have similar quality. 

\subsection{ProgGAN}

Taking into account the network structure and training algorithm previously discussed, we made one experiment regarding the ProgGAN model. To be able to verify the efficiency in training time we had to run an experiment E7 with similar amount of epochs to E6.  Therefore, we trained the model for $4950$ epochs ( $1650$ epochs for each of the networks ). 

Table \ref{tab:training_time_ProgGAN} depict the training time for E6 and E7. As we can see, the ProgGAN model took less $57$ hours to train than the WGAN model.

\begin{table}[htb]
\centering
\normalsize
    \caption{Training time for each experiment}
    \label{tab:training_time_ProgGAN}
{\footnotesize
    \begin{tabular}{ | l | l | l |}
    \hline 
     & E6 & E7  \\  \hline
     $64 \times 64$  & & 25h 3min    \\ \hline
     growth  &  & 74h 9min    \\  \hline 
     $128 \times 128$  & 217h & 71h 10min   \\ \hline
    TOTAL &  217h & 170h 23 min \\  \hline
    \end{tabular}
    }
\end{table}

\begin{figure}[h]
\centering
\includegraphics[width=0.5\textwidth]{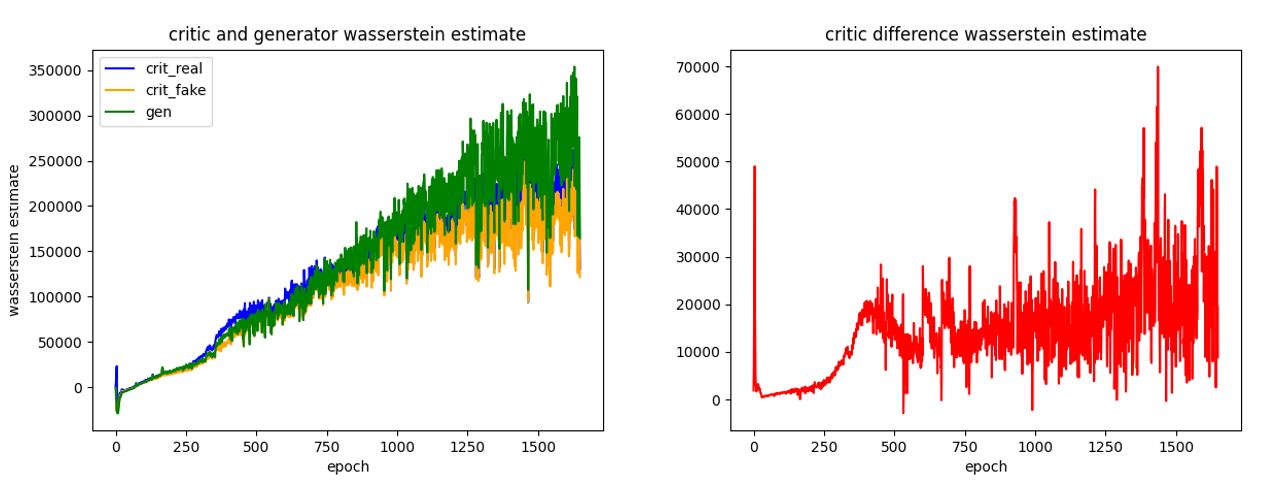}
\caption{Left: Wasserstein estimate for $C$ real and fake images and $G$ generated images during the training phase of the $64 \times 64$ model; Right: Difference between the $C$ Wasserstein estimate of the real images and the fake images}
\label{fig:ProgGAN_64x64}
\end{figure}
\begin{figure}[h!]\centering\includegraphics[width=0.5\textwidth]{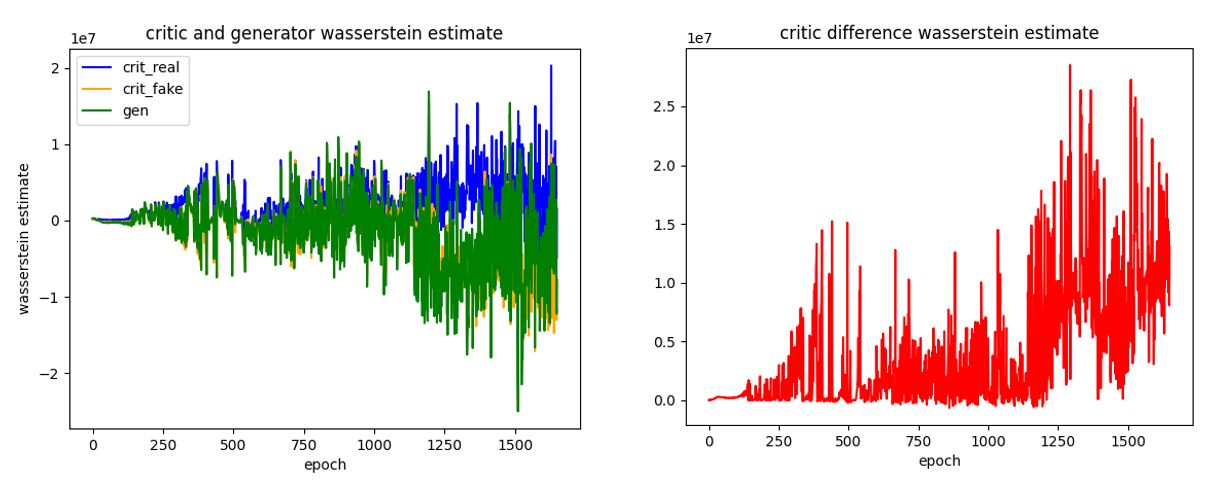}
\caption{Left: Wasserstein estimate for $C$ real and fake images and $G$ generated images during the training phase of the growth model; Right: Difference between the $C$ Wasserstein estimate of the real images and the fake images}
\label{fig:ProgGAN_growth}\end{figure}
\begin{figure}[h!]
\centering
\includegraphics[width=0.5\textwidth]{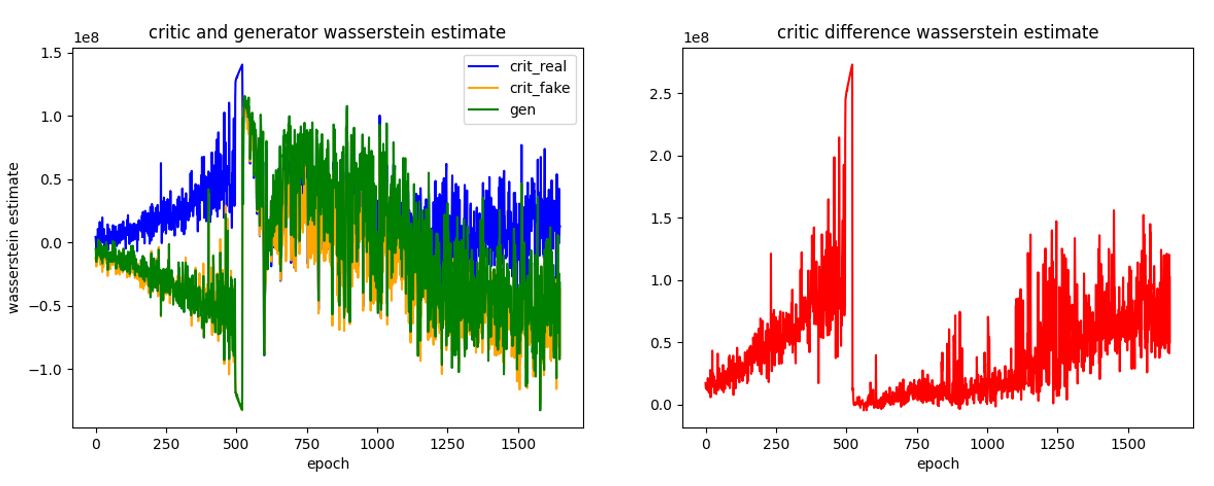}
\caption{Left: Wasserstein estimate for $C$ real and fake images and $G$ generated images during the training phase of the $128 \times 128$ model; Right: Difference between the $C$ Wasserstein estimate of the real images and the fake images}
\label{fig:ProgGAN_128x128}
\end{figure}

Fig \ref{fig:ProgGAN_64x64}, \ref{fig:ProgGAN_growth} and \ref{fig:ProgGAN_128x128} depict the Wasserstein estimate of both generators and critics for the $64 \times 64$, growth and $128 \times 128$ models, respectively.

Despite the figures showing an erratic and unstable movement of the Wasserstein estimate during the training phase, we can see that in all of the models, the estimates of the real and fake images were moving away from each other, hence the line of the difference of the Wasserstein estimate on the critic going up. 



\begin{figure}[h!]
\centering
\includegraphics[width=0.45\textwidth]{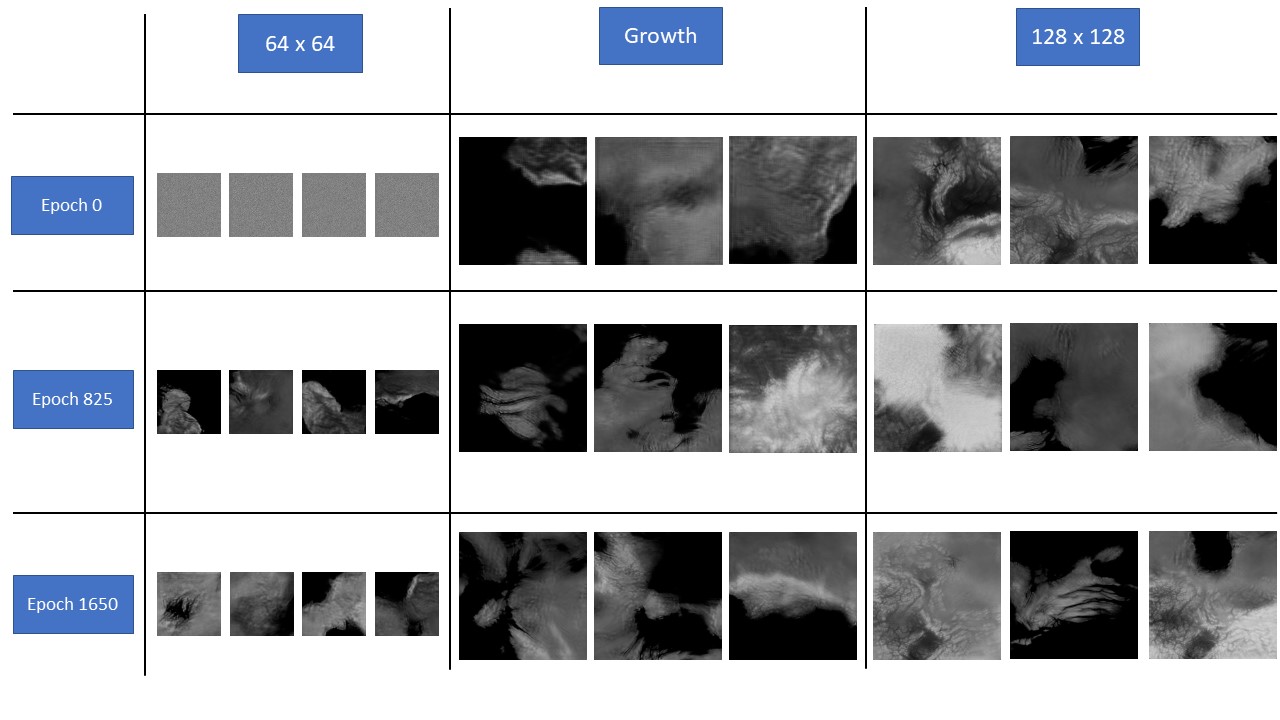}
\caption{Images generated by the ProgGAN's generator during the training epochs for the several models' networks in E7}
\label{fig:ProgGAN_images}
\end{figure}

Fig. \ref{fig:ProgGAN_images} shows some of the images generated through some training epochs of E7. We can see that the quality of the images from the $64 \times 64$ model to the growth model and to the $128 \times 128$ didn't deteriorate. We can even say that the quality improved, since there is more variety of complex structures present in the end of the training phase such as bays, peninsulas, mountain ranges near the shore and islands. 

\subsection{VAE + WGAN}

We did several experiments with the VAE + WGAN architecture, to analyze factors such as:
\begin{itemize}
    \item Number of epochs needed to train the VAE in order for it to recreate the dataset images with a reasonable level of detail;
    \item Generation of the random $z$ vector in order to be given as input to WGAN's generator;
    \item Comparison between training a WGAN with the decoder having already learned features from the Heightmap dataset and a WGAN without a pre-trained decoder.
\end{itemize}   
Taking into account these factors the experiments are described as follow:

\begin{itemize}
    \item E8: VAE trained for $1000$ epochs. Decoder with the weights already trained used as the WGAN's generator. 
    WGAN trained for $1000$ more epochs, using a normal distribution with $\mu = 0$ and $\sigma = 1$ for generating $z$;
    \item E9: VAE trained for $150$ epochs. 
    Decoder with the weights already trained used as the WGAN's generator. WGAN trained for $1000$ more epochs, using a normal distribution with $\mu = 0$ and $\sigma = 1$ for generating $z$;
    \item E10: Decoder, with the weights already trained in Experiment $2$, used as the WGAN's generator. WGAN trained for $1000$ more epochs, using the vectors $\mu$ and $\sigma$ learned from the VAE, for generating $z$. 
    \item E11: WGAN trained for $1150$ epochs to make a comparison between the models.
\end{itemize}

The training time for each experiment is denoted in Table \ref{tab:training_time_VAEGAN}. As we can see, the difference between E9 or E10, which corresponds to training VAE for $150$ epochs and training the WGAN for $1000$ epochs and E11  which corresponds to training the WGAN for $1150$ epochs isn't that big, saving at maximum $20$ minutes with the latter experiment. Fig. \ref{fig:VAE-WGAN_2_all_experiments_images} below depicts images, for the several experiments, generated by the WGAN after training.

\begin{table}[htb]
\centering
\normalsize
    \caption{Training time for each experiment}
    \label{tab:training_time_VAEGAN}
{\footnotesize
    \begin{tabular}{ | c | c | c | c |}
    \hline 
    Experiment	& VAE & WGAN & Total\\ \hline
    E8  & 8h 42 min & 7h 28 min & 16h 8 min  \\ \hline
    E9  & 1h 18 min & 7h 30 min & 8h 48 min \\ \hline 
    E10 & 1h 18 min & 7h 20 min & 8h 38 min  \\ \hline
    E11 & & 8h 24 min & 8h 24 min \\ \hline
    \end{tabular}
    }
\end{table}

\begin{figure}[h!]
\centering
\includegraphics[width=0.5\textwidth]{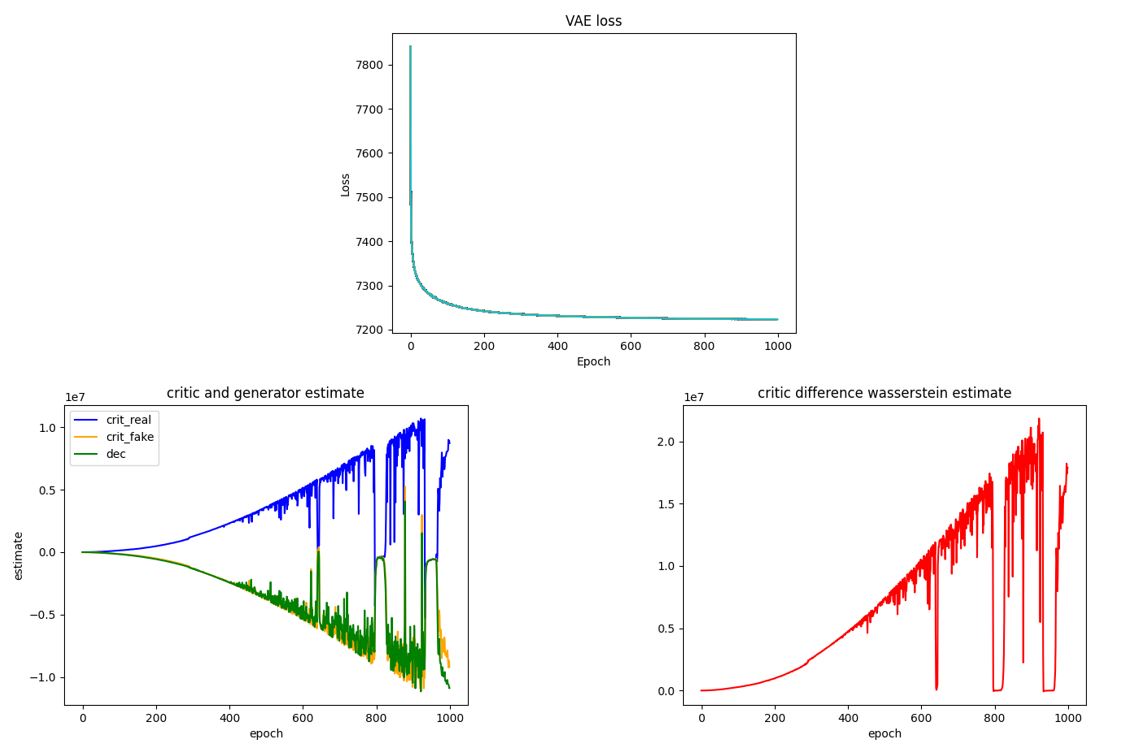}
\caption{E8: Top: VAE' loss during the training phase;
Left: Wasserstein estimate for $C$ real and fake images and decoder's generated images during the training phase; Right: Difference between the $C$ Wasserstein estimate of the real images and the fake images}
\label{fig:VAE-WGAN_1}
\end{figure}

Regarding E8, the loss and estimates are shown in Fig. \ref{fig:VAE-WGAN_1}. The VAE's training went well as expected, with its loss rapidly decreasing in the initial epochs, but stagnating around $7250$ towards the end of the training. The WGAN's training also went smoothly, despite some moments between the epochs $800$ and $1000$ where all the estimates suddenly got close to each other.

\begin{figure}[h!]
\centering
\includegraphics[width=0.5\textwidth]{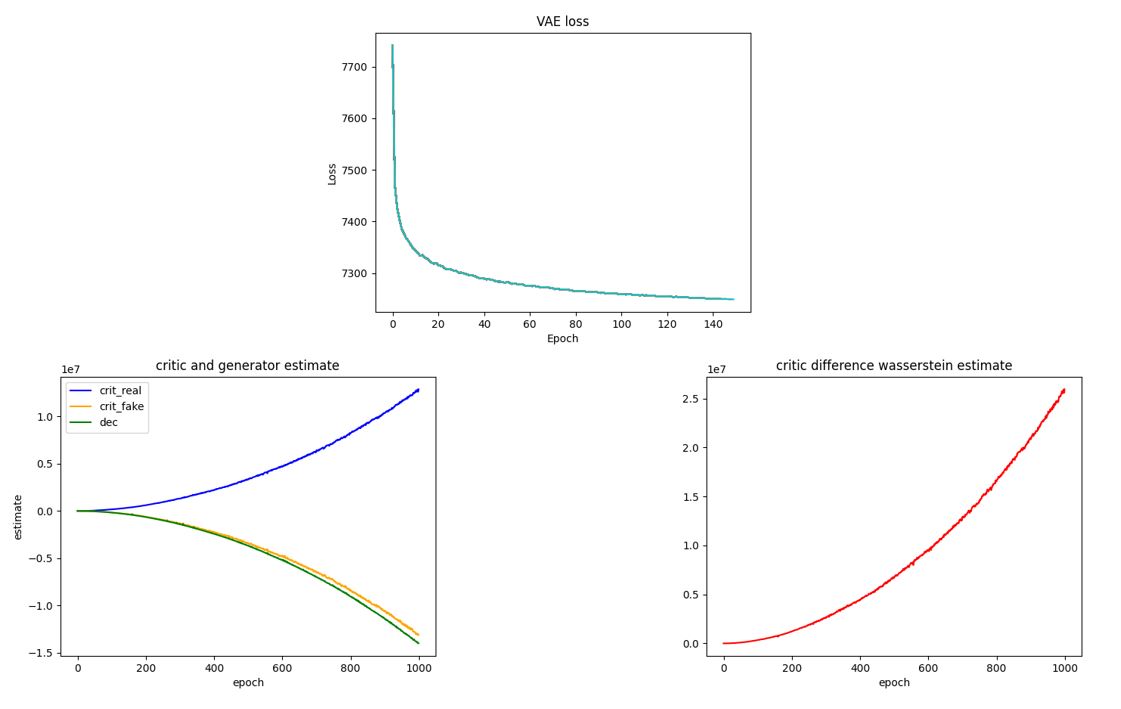}
\caption{E9: Top: VAE' loss during the training phase;
Left: Wasserstein estimate for $C$ real and fake images and decoder's generated images during the training phase; Right: Difference between the $C$ Wasserstein estimate of the real images and the fake images}
\label{fig:VAE-WGAN_2}
\end{figure}

\begin{figure}[h!]
\centering
\includegraphics[width=0.45\textwidth]{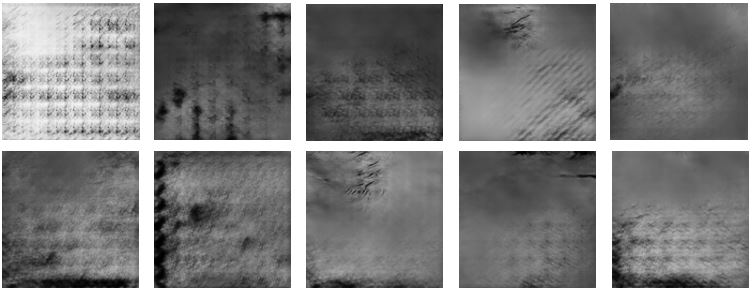}
\caption{Images with some strange artifacts generated by WGAN during the epochs for E10}
\label{fig:VAE-WGAN_2_images_artifacts}
\end{figure}

Since the stagnation of VAE's loss happened early in its training phase we thought we could have trained it for a lower amount of epochs and still be able to generate images with good quality. Taking that into account, in E9, as shown in Fig. \ref{fig:VAE-WGAN_2}, 
we trained the VAE for only $150$ epochs. In this experiment, the WGAN training went really smoothly, as can be seen by the Wasserstein estimate graphic and the images generated. However, in this experiment, there were images, generated throughout the training process, that presented some strange artifacts as seen in Fig. \ref{fig:VAE-WGAN_2_images_artifacts}. Even after the training was complete, some images with those kind of artifacts were generated.

\begin{figure}[h!]
\centering
\includegraphics[width=0.5\textwidth]{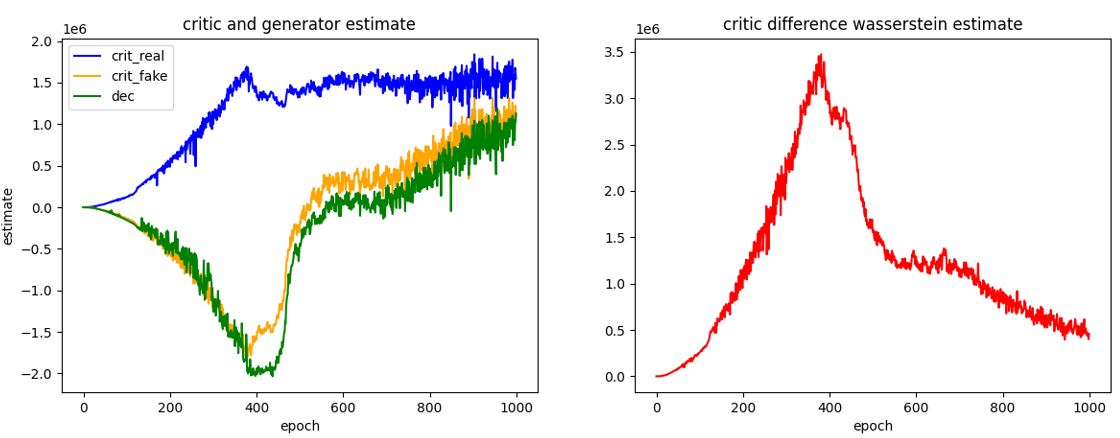}
\caption{E10: Left: Wasserstein estimate for $C$ real and fake images and decoder's generated images during the training phase; Right: Difference between the $C$ Wasserstein estimate of the real images and the fake images}
\label{fig:VAE-WGAN_3}
\end{figure}

In E10 we took the weights of the decoder trained in the VAE of the previous experiment and trained the WGAN for $1000$ epochs again. However, we are using the vectors $\mu$ and $\sigma$ learned from the VAE to generate our vectors $z$, since they contain features of the Heightmap dataset. Fig. \ref{fig:VAE-WGAN_3} and \ref{fig:VAE-WGAN_2_all_experiments_images} depict the results of E10. The graphic of the Wasserstein estimate is not what we expected it to be. From epoch $400$, the images generated by the decoder managed to fool the critic into thinking they were becoming more real throughout the remaining epochs, given that the Wasserstein estimate of the fake images went up after that epoch. 

\begin{figure}[h!]
\centering
\includegraphics[width=0.5\textwidth]{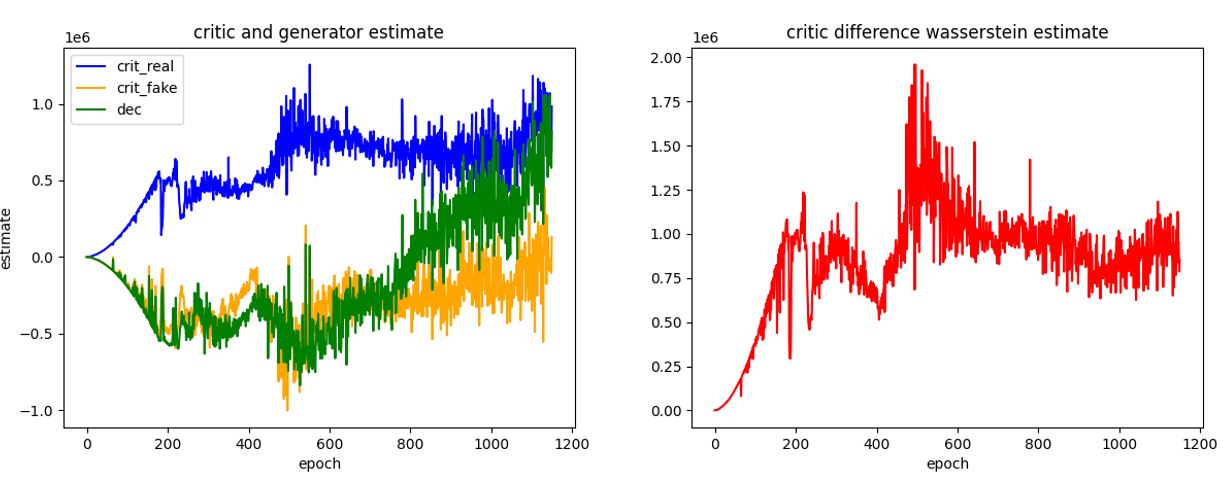}
\caption{E11: Left: Wasserstein estimate for $C$ real and fake images and decoder's generated images during the training phase; Right: Difference between the $C$ Wasserstein estimate of the real images and the fake images}
\label{fig:VAE-WGAN_4}
\end{figure}

As the last experiment, we wanted to just train the WGAN for $1150$ epochs, in order to make a proper comparison between the other experiments. Fig. \ref{fig:VAE-WGAN_4} and \ref{fig:VAE-WGAN_2_all_experiments_images} depict the results. 




\begin{figure}[h!]
\centering
\includegraphics[width=0.45\textwidth]{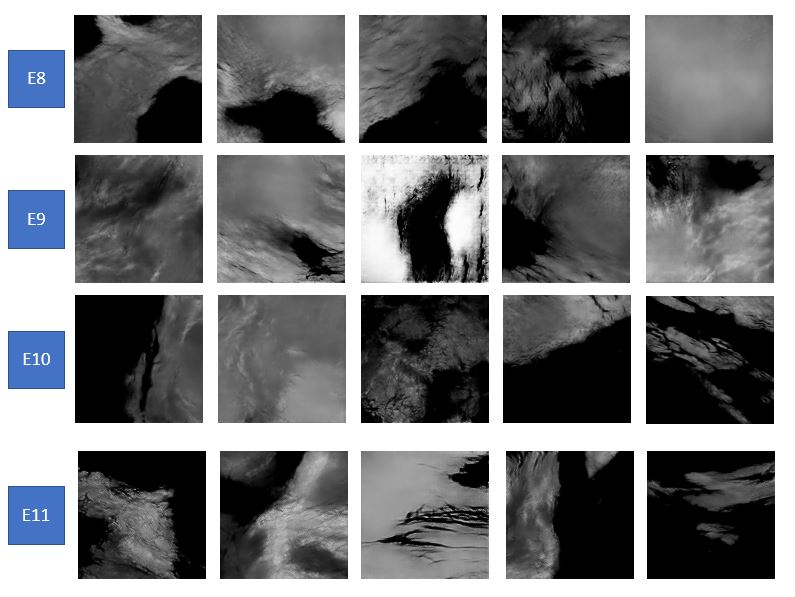}
\caption{Images generated by WGAN after training, in the several experiments}
\label{fig:VAE-WGAN_2_all_experiments_images}
\end{figure}

\subsection*{Discussion}
The results obtained with the DCGAN were unsatisfying, as we were already expecting, due to several problems with this architecture already mentioned in the literature. The WGAN proved to be a good option since throughout the whole training phase the generator always had enough information provided by the critic to improve itself, thus ending up generating heightmaps with good quality. With the ProgGAN we saw that we could achieve the same results as in the WGAN with less training time. Curiously, the Wasserstein estimates did not follow the expected pattern. With the VAE + WGAN model, although it seemed to be a promising model in paper, we didn't get quite the results we expected.

\section{Conclusions}

In this work, our purpose was to explore the GANs' capabilities to generate images with high resolution and quality to try to create maps that look realistic and appealing for players, in a visual and even strategic way. That could not be done with the traditional approaches of PCG. Instead of just focusing on one  model of the current state of the art of GANs we decided to explore several ones and test if they could indeed be used to generate such maps. We ended up exploring four  models: DCGAN, WGAN, ProgGAN and VAE + WGAN, each one with its upsides and downsides. The DCGAN and VAE + WGAN models provided results with poor quality while the WGAN and ProgGAN models provided the best results, with the latter one being more efficient in terms of training time. Despite the poor results of the VAE + WGAN model, we think that this last model is one that should be further studied because its concept is promising. 

Given the server memory limitations we were only able to work with images of $128 \times 128$ resolution while the ideal scenario would be to work with an higher resolution such as $1024 \times 1024$. Further individual studies of each of these  models or the ability to be able to expand these networks to images with higher resolution, given the proper hardware, are some examples of future work that could be explored and studied regarding this topic.


\section*{Acknowledgments}

This work was supported by national funds through FCT, Fundação para a Ciência e a Tecnologia, under projects  UIDB/04326/2020, UIDB/50021/2020, UIDP/04326/2020 and LA/P/0101/2020.

\printbibliography


\begin{acronym}[H.264/SVC]
    \acro{ANN}{Artificial Neural Networks}
    \acro{BN}{Batch Normalization}
    \acro{CNN}{Convolutional Neural Networks}
    \acro{DCGAN}{Deep Convolutional Generative Adversarial Network}
    \acro{DEM}{Digital Elevation Model}
    \acro{GAN}{Generative Adversarial Networks}
    \acro{IST}{Instituto Superior Tecnico}
	\acro{PCG}{Procedural Content Generation}
	\acro{ProgGAN}{Progressive Growing Generative Adversarial Network}
	\acro{ReLU}{Rectified Linear Unit}
	\acro{SRTM}{Shuttle Radar Topography Mission}
	\acro{VAE}{Variational Autoencoder}
	\acro{WGAN}{Wasserstein Generative Adversarial Network}
\end{acronym}


\section*{Appendix A} 

In this appendix we show the detailed architecture of the implemented models and the hyper-parameters used, to allow for reproducibility of the results. For the DCGAN network, we used the Adam Optimizer with learning rate 0.0002 and $\beta_1$ = 0.5.
For the WGAN and ProgGAN network, we used the RMSProp optimizer with learning rate 0.0005. For the VAE + WGAN model we trained the VAE and then the whole model, using the RMSProp optimizer with learning rate 0.0003 and 0.0005, respectively.

\begin{table}[htbp]
\caption{Generator structure of the DCGAN model used}
\begin{center}
\begin{tabular}{|lcll|}

\hline
Layer name & 
Act. &
Input shape &
Output shape
\\
\hline 

Dense & - &  $100  \times 1 \times 1$ & $65536 \times 1 \times 1$ \\
Reshape & - & $65536 \times 1 \times 1$ & $1024 \times 8 \times 8$ \\

\hline
Deconv & -  & $1024 \times 8 \times 8$ & $256 \times 16 \times 16$ \\
BatchNorm & - & $256 \times 16 \times 16$ & $256 \times 16 \times 16$ \\
LeakyReLU & LeakyReLU & $256 \times 16 \times 16$ & $256 \times 16 \times 16$ \\

Dropout & - & $256 \times 16 \times 16$ & $256 \times 16 \times 16$ \\

\hline
Deconv & - & $256 \times 16 \times 16$ & $128 \times 32 \times 32$ \\
BatchNorm & - & $128 \times 32 \times 32$ & $128 \times 32 \times 32$ \\
LeakyReLU & LeakyReLU & $128 \times 32 \times 32$ & $128 \times 32 \times 32$ \\

Dropout & - & $128 \times 32 \times 32$ & $128 \times 32 \times 32$ \\

\hline
Deconv & - & $128 \times 32 \times 32$ & $64 \times 64 \times 64$ \\
BatchNorm & - & $64 \times 64 \times 64$ & $64 \times 64 \times 64$ \\
LeakyReLU & LeakyReLU & $64 \times 64 \times 64$ & $64 \times 64 \times 64$ \\

Dropout & - & $64 \times 64 \times 64$ & $64 \times 64 \times 64$ \\
\hline
Deconv & - & $64 \times 64 \times 64$ & $32 \times 128 \times 128$ \\
BatchNorm & - & $32 \times 128 \times 128$ & $32 \times 128 \times 128$ \\
LeakyReLU & LeakyReLU & $32 \times 128 \times 128$ & $32 \times 128 \times 128$ \\

Dropout & - & $32 \times 128 \times 128$ & $32 \times 128 \times 128$ \\

\hline
Deconv & tanh & $32 \times 128 \times 128$ & $1 \times 128 \times 128$ \\
\hline
\end{tabular}
\end{center}
\label{tab:DCGAN_G} 
\end{table}

\begin{table}[htbp]
\caption{Discriminator structure of the DCGAN model used}
\begin{center}
\begin{tabular}{|lcll|}

\hline
Layer name & 
Act. &
Input shape &
Output shape
\\
\hline 

Conv & - & $1 \times  128 \times 128$ & $1 \times 128 \times 128$ \\
LeakyReLU & LeakyReLU & $1 \times 128 \times 128$ & $1 \times 128 \times 128$ \\

\hline
Conv & - &  $1 \times 128 \times 128$ & $32 \times 64 \times 64$ \\
BatchNorm & - & $32 \times 64 \times 64$ & $32 \times 64 \times 64$ \\
LeakyReLU & LeakyReLU &  $32  \times 64 \times 64$ & $32 \times 64 \times 64$ \\

\hline
Conv & - &  $32 \times 64 \times 64$ & $64 \times 32 \times 32$ \\
BatchNorm & - & $64 \times 32 \times 32$ & $64 \times 32 \times 32$ \\
LeakyReLU & LeakyReLU & $64 \times 32 \times 32$ & $64 \times 32 \times 32$ \\

\hline
Conv & - & $64 \times 32 \times 32$ & $128 \times 16 \times 16$ \\
BatchNorm & - & $128 \times 16 \times 16$ & $128 \times 16 \times 16$ \\
LeakyReLU & LeakyReLU & $128 \times 16 \times 16$ & $128 \times 16 \times 16$ \\

\hline
Conv & - &  $128 \times 16 \times 16$ & $256 \times 8 \times 8$ \\
BatchNorm & - & $256 \times 8 \times 8$ & $256 \times 8 \times 8$ \\
LeakyReLU & LeakyReLU & $256 \times 8 \times 8$ & $256 \times 8 \times 8$ \\

\hline
Flatten & - & $256 \times 8 \times 8$ & $16384 \times 1 \times 1$ \\
Dense & sigmoid & $16384 \times 1 \times 1$ & $1 \times 1 \times 1$ \\
\hline
\end{tabular}
\end{center}
\label{tab:DCGAN_D} 
\end{table}

\begin{table}[htbp]
\label{Table-blocks}
\caption{Blocks of layers used in the ProgGAN network's structures}
\begin{center}
\begin{tabular}{|c|lcl|}

\hline
&
Layer name & 
Act. &
Output shape
\\
\hline 

\hline
 & Deconv & - &  
$128 \times 64 \times 64$  \\
& BatchNorm & - &  
$128 \times 64 \times 64$ \\
& LeakyReLU & LeakyReLU & 
$128 \times 64 \times 64$ \\

\cline{2-4}
$DECONV\_1$ & Deconv & - & 
$128 \times 64 \times 64$ \\
& BatchNorm & - &  
$128 \times 64 \times 64$  \\
& LeakyReLU & LeakyReLU & 
$128 \times 64 \times 64$ \\

\hline

& UpSampling & - & 
$128 \times 128 \times 128$ \\
& Deconv & - &  
$64 \times 128 \times 128$  \\
& BatchNorm & - & 
$64 \times 128 \times 128$ \\
$DECONV\_2$& LeakyReLU & LeakyReLU & 
$64 \times 128 \times 128$ \\

\cline{2-4}
 & Deconv & - & 
$64 \times 128 \times 128$ \\
& BatchNorm & - & 
$64 \times 128 \times 128$  \\
& LeakyReLU & LeakyReLU & 
$64 \times 128 \times 128$ \\

\hline

& Conv & - & 
$64 \times 128 \times 128$  \\
& BatchNorm & - & 
$64 \times 128 \times 128$  \\
& LeakyReLU & LeakyReLU & 
$64 \times 128 \times 128$  \\

\cline{2-4}
$CONV\_1$ & Conv & - & 
$128 \times 128 \times 128$  \\
& BatchNorm & - & 
$128 \times 128 \times 128$  \\
& LeakyReLU & LeakyReLU & 
$128 \times 128 \times 128$  \\
& Downsample & - & 
$128 \times 64 \times 64$  \\

\hline

& Conv & - &  
$128 \times 64 \times 64$  \\
& BatchNorm & - & 
$128 \times 64 \times 64$  \\
$CONV\_2$& LeakyReLU & LeakyReLU & 
$128 \times 64 \times 64$  \\

\cline{2-4}
& Conv & - & 
$128 \times 64 \times 64$ \\
& BatchNorm & - & 
$128 \times 64 \times 64$  \\
& LeakyReLU & LeakyReLU & 
$128 \times 64 \times 64$  \\

\hline

\end{tabular}
\end{center}
\label{tab:block_layers}
\end{table}

\begin{table}[htbp]
\caption{Generator structure of the $64 \times 64$ network}
\begin{center}
\begin{tabular}{|lcll|}

\hline
Layer name & 
Act. &
Input shape &
Output shape
\\
\hline

Dense & - & $100  \times 1 \times 1$ & $262144 \times 1 \times 1$ \\
Reshape & - & $262144 \times 1 \times 1$ & $64 \times 64 \times 64$ \\

\hline
  & & &  \\
  \multicolumn{4}{|c|}{$DECONV\_1$}   \\
  & & & \\

\hline
 Deconv & Tanh & $128 \times 64 \times 64$ & $1 \times 64 \times 64$ \\
 
\hline

\end{tabular}
\end{center}
\label{tab:PGAN_G_64x64}
\end{table}

\begin{table}[htp!]
\caption{Critic structure of the $64 \times 64$ network}
\begin{center}
\begin{tabular}{|lcll|}

\hline
Layer name & 
Act. &
Input shape &
Output shape
\\
\hline

Conv & - & $1 \times  64 \times 64$ & $128 \times 64 \times 64$ \\
LeakyReLU & LeakyReLU & $128 \times 64 \times 64$ & $128 \times 64 \times 64$ \\

\hline
  & & &  \\
  \multicolumn{4}{|c|}{$CONV\_2$}   \\
  & & & \\

\hline
Flatten & - &  $128 \times 64 \times 64$ & $524288 \times 1 \times 1$ \\
Dense & linear & $524288 \times 1 \times 1$ & $1 \times 1 \times 1$ \\
\hline

\end{tabular}
\end{center}
\label{tab:PGAN_C_64x64}
\end{table}

\begin{table}[htp!]
\caption{Generator structure of the $128 \times 128$ network}
\begin{center}
\begin{tabular}{|lcll|}

\hline
Layer name & 
Act. &
Input shape &
Output shape
\\
\hline

Dense & - & $100  \times 1 \times 1$ & $262144 \times 1 \times 1$ \\
Reshape & - & $262144 \times 1 \times 1$ & $64 \times 64 \times 64$ \\

\hline
  & & &  \\
  \multicolumn{4}{|c|}{$DECONV\_1$}   \\
  & & & \\

\hline
  & & &  \\
  \multicolumn{4}{|c|}{$DECONV\_2$}   \\
  & & & \\

\hline
 Deconv & Tanh & $64 \times 128 \times 128$ & $1 \times 128 \times 128$ \\
\hline

\end{tabular}
\end{center}
\label{tab:PGAN_G_128x128} 
\end{table}

\begin{table}[ht!]
\caption{Critic structure of the $128 \times 128$ network}
\begin{center}
\begin{tabular}{|lcll|}

\hline
Layer name & 
Act. &
Input shape &
Output shape
\\
\hline

Conv & - & $ 1 \times  128 \times 128$ & $64 \times 128 \times 128$ \\
LeakyReLU & LeakyReLU & $64 \times 128 \times 128$ & $64 \times 128 \times 128$ \\

\hline
  & & &  \\
  \multicolumn{4}{|c|}{$CONV\_1$}   \\
  & & & \\
 
\hline
  & & &  \\
  \multicolumn{4}{|c|}{$CONV\_2$}   \\
  & & & \\

\hline
 Flatten & - & $128 \times 64 \times 64$ & $524288 \times 1 \times 1$ \\
 Dense & linear & $524288 \times 1 \times 1$ & $1 \times 1 \times 1$ \\

\hline
\end{tabular}
\end{center}
\label{tab:PGAN_C_128x128}
\end{table}

\begin{table}[htbp]
\caption{Encoder structure of the VAE + WGAN model used} \label{tab:VAEGAN_Encoder}
\begin{center}
\begin{tabular}{|lcll|}

\hline
Layer name & 
Act. &
Input shape &
Output shape
\\
\hline 

Conv & - &  $1 \times  128 \times 128$ & $16 \times 64 \times 64$ \\
BatchNorm & - &  $16 \times 64 \times 64$ & $16 \times 64 \times 64$ \\
ReLU & ReLU & $16 \times 64 \times 64$ & $16 \times 64 \times 64$ \\

\hline
Conv & - & $16 \times 64 \times 64$ & $32 \times 32 \times 32$ \\
BatchNorm & - &  $32 \times 32 \times 32$ & $32 \times 32 \times 32$ \\
ReLU & ReLU &  $32  \times 32 \times 32$ & $32 \times 32 \times 32$ \\

\hline
Conv & - &  $32 \times 32 \times 32$ & $64 \times 16 \times 16$ \\
BatchNorm & - &  $64 \times 16 \times 16$ & $64 \times 16 \times 16$ \\
ReLU & ReLU &  $64 \times 16 \times 16$ & $64 \times 16 \times 16$ \\

\hline
Conv & - &  $64 \times 16 \times 16$ & $128 \times 8 \times 8$ \\
BatchNorm & - & $128 \times 8 \times 8$ & $128 \times 8 \times 8$ \\
ReLU & ReLU & $128 \times 8 \times 8$ & $128 \times 8 \times 8$ \\

\hline
Flatten & - &  $128 \times 8 \times 8$ & $8192 \times 1 \times 1$ \\
Dense & - &  $8192 \times 1 \times 1$ & $1024 \times 1 \times 1$ \\
BatchNorm & - &  $1024 \times 1 \times 1$ & $1024 \times 1 \times 1$ \\
ReLU & ReLU &  $1024 \times 1 \times 1$ & $1024 \times 1 \times 1$ \\
\hline

\hline
$\mu$ (Dense) & - &  $1024 \times 1 \times 1$ & $512 \times 1 \times 1$ \\ 
$\sigma$ (Dense) & - &  $1024 \times 1 \times 1$ & $512 \times 1 \times 1$ \\ 
\hline

\end{tabular}
\end{center}

\end{table}

\end{document}